\documentclass[conference]{IEEEtran}
\IEEEoverridecommandlockouts
\usepackage{cite}
\usepackage{amsmath,amssymb,amsfonts}
\usepackage{algorithmic}
\usepackage{graphicx}
\usepackage{textcomp}
\usepackage{xcolor}

\usepackage{subfigure}
\usepackage{booktabs}
\usepackage{array}
\usepackage{mathtools}
\usepackage{amsthm}
\usepackage[capitalize,noabbrev]{cleveref}

\def\BibTeX{{\rm B\kern-.05em{\sc i\kern-.025em b}\kern-.08em
    T\kern-.1667em\lower.7ex\hbox{E}\kern-.125emX}}
\begin{document}

\title{Real-Time Assessment of Bystander's Situation Awareness in Drone-Assisted First Aid}

\author{\IEEEauthorblockN{Shen Chang\textsuperscript{1}, 
Renran Tian\textsuperscript{2}, 
Nicole Adams\textsuperscript{3}, 
Nan Kong\textsuperscript{1,4}}
\IEEEauthorblockA{\textsuperscript{1}Weldon School of Biomedical Engineering,
Purdue University, West Lafayette, Indiana, USA\\
chang887@purdue.edu, nkong@purdue.edu}
\IEEEauthorblockA{\textsuperscript{2}Edward P. Fitts Department of Industrial and Systems Engineering, 
North Carolina State University, Raleigh, NC, USA\\
rtian2@ncsu.edu}
\IEEEauthorblockA{\textsuperscript{3}Regenstrief Center for Healthcare Engineering, 
Purdue University, West Lafayette, Indiana, USA\\
adams417@purdue.edu}
\IEEEauthorblockA{\textsuperscript{4}Edwardson School of Industrial Engineering,
Purdue University, West Lafayette, Indiana, USA\\}
}

\maketitle

\begin{abstract}
Rapid naloxone delivery via drones offers a promising solution for responding to opioid overdose emergencies (OOEs), by extending lifesaving interventions to medically untrained bystanders before emergency medical services (EMS) arrive. Recognizing the critical role of bystander's situational awareness (SA) in human-autonomy teaming (HAT), we address a key research gap in real-time SA assessment by introducing the Drone-Assisted Naloxone Delivery Simulation Dataset (DANDSD). This pioneering dataset captures HAT during simulated OOEs, where college students without medical training act as bystanders tasked with administering intranasal naloxone (Narcan™) to a mock overdose victim. Leveraging this dataset, we propose a video-based real-time SA assessment framework that utilizes graph embeddings and transformer models to assess bystander SA in real time. Our approach integrates visual perception and comprehension cues—such as geometric, kinematic, and interaction graph features—and achieves high-performance SA prediction. It also demonstrates strong temporal segmentation accuracy, outperforming the FINCH baseline by 9\% in Mean over Frames (MoF) and 5\% in Intersection over Union (IoU). This work supports the development of adaptive drone systems capable of guiding bystanders effectively, ultimately improving emergency response outcomes and saving lives. The dataset and source code are publicly available at https://github.com/chang887/DANDSD, enabling continued research in this vital area.
\end{abstract}


\section{Introduction}\label{sec:intro}
Effective situational awareness (SA) is the cornerstone of successful first aid in out-of-hospital medical emergencies (OHME), guiding bystanders and first responders to make informed, life-saving decisions. It enables them to perceive, comprehend, and project the status of their environment and the individuals involved \cite{endsley1995toward}. However, achieving and maintaining SA can be particularly challenging in time-sensitive OHME such as stroke, cardiac arrest, and opioid overdose. Research indicates that the odds of survival from out-of-hospital cardiac arrest (OHCA) decrease by 7–17\% for every minute without treatment \cite{stoesser2021moderating}. Likewise, substance overdose incidents, road traffic accidents, and maternal health issues require immediate attention to prevent fatalities. In these situations, delayed response times and limited access to emergency medical services (EMS) can significantly impair bystanders' ability to gather and process information, leading to suboptimal patient outcomes \cite{mell2017emergency}.

In response to these challenges, the increasing use of unmanned aerial vehicles (UAVs), also known as drones, for delivering life-saving interventions, such as automated external defibrillators (AEDs) and naloxone, promises faster response times and improved prehospital patient outcomes \cite{boutilier2017optimizing, ornato2020feasibility}. However, the effectiveness of these aids heavily depends on the collaboration between the drone and the bystander, often the first to recognize the emergency and initiate the 9-1-1 call. Studies have demonstrated that bystanders frequently underperform in first aid situations \cite{wissenberg2013association,ritter1985effect}, revealing a critical gap between technological advancements and human performance.

In this context, modern artificial intelligence (AI) is well positioned to play a key role by enhancing bystanders' SA and providing real-time guidance to improve decision-making processes. AI systems could analyze real-time data collected by drones, evaluate the bystander's current level of SA, and provide effective, context-specific instructions and operational demonstrations. For instance, AI could assist a bystander by verifying proper electrode pad placement, confirming scene safety prior to shock delivery, and providing structured guidance for the administration of naloxone nasal spray during an opioid overdose. By adapting to the bystander's evolving SA, AI could help ensure timely and appropriate actions, leading to more successful rescues.

A key advantage of leveraging AI to enhance human SA and decision-making is its ability of scene understanding in real-time. Traditional SA assessment techniques rely on subjective measurements and post-hoc evaluations, making them unsuitable for real-time applications \cite{dahn2018situation}. These methods do not accommodate the unique capability of AI systems, such as processing large volumes of video stream data and managing inherent uncertainties in real time. Alternative approaches, including physiological measurements and computational models, have been explored. Physiological measurements, such as brain activity monitoring, show promise but struggle to establish robust links between physiological data and mental performance \cite{zhang2023physiological}. Existing computational models provide a more precise evaluation of SA but face challenges in adapting to dynamic real-time scenarios and integrating human input effectively \cite{dahn2018situation,muller2023self}. These challenges are pronounced in OHME, where quick effective decisions are essential.

Given the complexity of OHME scenarios, there is a pressing need for innovative machine learning (ML) methodology that is capable of real-time SA assessment to enable adaptive decision-making. These techniques must address the unique challenges of bystander-drone cooperation by accurately assessing the real-time SA of first-aid bystanders and identifying temporal-dynamic changes in the situation. With such SA assessment, adaptive AI systems can be effectively developed, ultimately saving lives and improving patient outcomes.

This study aims to enhance SA in OHME through the integration of SA-focused bystander-drone interaction data analysis and imitation learning. Our primary objective is to develop an AI framework that leverages visual features to emulate SA assessment of medical experts. We propose a Transformer-based AI framework for the assessment in a simulated drone-assisted opioid overdose emergency. We expect to achieve the following three main contributions:
\begin{itemize}
    \item \textbf{First-of-its-kind Bystander-Drone Interaction Dataset:} The collection of bystander-drone interaction data during simulated OHME marks a significant milestone, as it's the first dataset from the observer's perspective. This dataset is annotated based on observer-rated SA, integrating perception, comprehension, and projection. The annotation process includes delineating event boundaries and formulating a single-scale SA metric, ensuring precision for AI model training.

    \item \textbf{Novel SA Assessment Framework:} We pioneer an AI framework that simplifies the prediction of human SA into a classification approach. This system integrates visual features with a transformer architecture and uses compositional learning to combine graph embeddings. Our framework captures complex spatiotemporal relationships among people, drones, and environmental factors, enhancing the understanding of dynamic environments.

    \item \textbf{Latent SA Labels Enhancing Temporal Segmentation Interpretation:} We connected SA evidence with temporal segmentation tasks, advancing the interpretation of segmentation results.  Tailored for OHME and complex human-autonomy teaming, our framework provides real-time feedback on SA-level fluctuations. This demonstrates its ability to identify event transitions and strengthens trust in AI capabilities.
\end{itemize}

\begin{figure*}[htb!]
\vskip 0.1in
\begin{center}
\centerline{\includegraphics[width=0.7\pdfpagewidth]{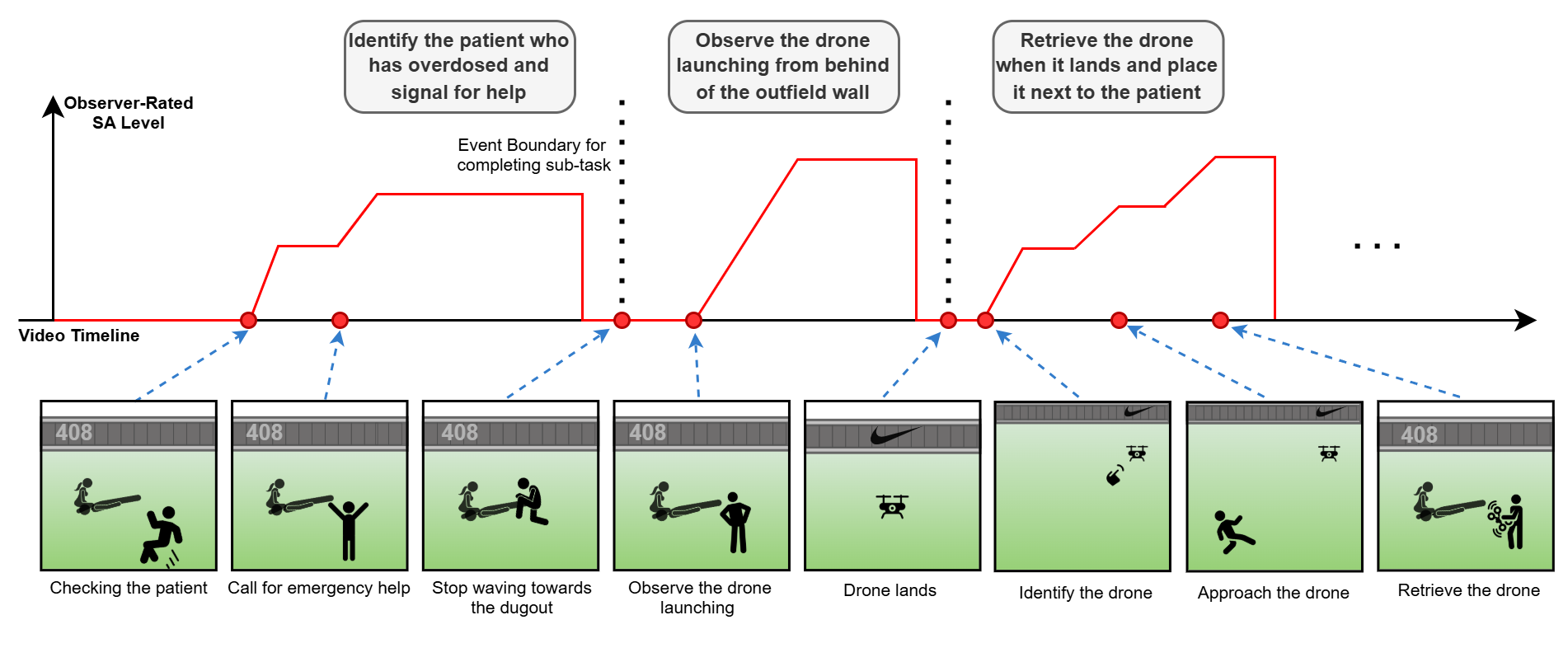}}
\vskip -0.1 in
\caption{Observer-rated SA changes during bystander-drone interaction using single-scale measurement.}
\label{fig:data}
\end{center}
\vskip -0.3 in
\end{figure*}

\section{Related Works}\label{sec:review}
\subsection{Traditional Situational Awareness (SA) Studies} 

SA assessment has garnered significant interest across domains, aiming to comprehend how individuals perceive, comprehend, and project information in complex environments. Numerous methods have emerged for assessing SA, which are broadly categorized into direct and indirect methods \cite{endsley1995toward}. Direct methods explicitly assess an individual's level of situational awareness (SA), using techniques such as the SA Global Assessment Technique (SAGAT) \cite{endsley1995toward}, post-test questionnaires evaluating situational knowledge \cite{durso1998situation}, and self-rating tools like the SA Rating Technique (SART) \cite{selcon1990evaluation}. Indirect methods, in contrast, assess an individual's level of SA based on performance outcomes or observable behaviors. Examples include behavioral marker systems, like the SA Behavioral Rating Scale (SABARS) \cite{matthews2002assessing} and the SA Linked Indicators Adapted to Novel Tasks (SALIANT) \cite{muniz1998methodology}, where trained observers rate participants on predefined behaviors believed to reflect SA. Performance outcome measures also yield indirect methods, inferring SA from task performance relative to some standard, such as the SA Probe Technique (SA-PT) \cite{jones2004use}. Recently, researchers have explored physiological measures, including eye-tracking data \cite{zhou2021using} and electroencephalography (EEG) \cite{fernandez2019encephalographic,zhang2023physiological}, to assess SA in real time. These approaches show promise for providing continuous, objective measures of SA without interrupting the task.

\subsection{AI-Assisted SA Assessment}

With advances in AI, SA research has expanded to new domains like autonomous ships \cite{thombre2020sensors} and vehicles \cite{zhou2021using}. Modern AI systems often rely on knowledge graphs and machine learning for SA computation. These learned representations support various tasks, including relevancy computation, similarity search, anomaly detection, prediction, and decision-making \cite{jiang2020improving}. However, many existing methods assume AI agents have complete knowledge of the situation, which is often not the case in dynamic environments \cite{jiang2020improving,dahn2018situation}. Additionally, coordinating multiple AI agents introduces complexities in data sharing and model integration, requiring novel definitions and frameworks for SA \cite{dahn2018situation}.

Effective measurement and improvement of SA in AI systems—especially in real-time applications—remain challenging. There is a need for new approaches that offer comprehensive metrics for assessing SA in real-world settings. We highlight the efficacy of observer-rated SA, a widely adopted indirect method used in medical applications \cite{cooper2014measuring}. Tools such as TEAM for resuscitation and patient deterioration \cite{cooper2010rating}, ANTS for anesthetic contexts \cite{fletcher2004rating} and intensive care units \cite{reader2006non}, and NOTSS for surgeons \cite{yule2006development}, leverage observer-rated SA. These tools involve experts observing individual or team performances and rating SA using predetermined scales. Our study extends individual SA measurement to three levels and emphasizes how analyzing event transitions can enhance SA assessment accuracy.

\subsection{Temporal Segmentation}
 
Temporal Segmentation (TS) has seen significant advancements in recent years, with approaches ranging from fully supervised to weakly supervised and unsupervised methods \cite{ding2023temporal}. Current state-of-the-art techniques leverage deep learning architectures such as Temporal Convolutional Networks (TCNs) \cite{lea2017temporal} and Transformers \cite{ulhaq2022vision} to enhance frame-wise representations and temporal modeling. Through recent breakthroughs in AI, researchers have started to focus on the interpretability of models. However, previous interpretable TS research has primarily focused on spatial attention mechanisms to identify important visual regions \cite{sun2022human}. Despite these advances, the integration of action segmentation with human SA evidence remains an underexplored area in the field. 


\section{Methodology}\label{sec:method}

Our previous simulation study established time baselines and measured experiences associated with bystander administration of drone-delivered naloxone in emergencies \cite{adams2023untrained}. Based on 17 simulated trials, we collected the Drone-Assisted Naloxone Delivery Simulation Dataset (DANDSD), comprising 11 continuous, uninterrupted videos that fully capture bystanders' actions and behaviors while administering drone-delivered naloxone to overdose patients (mannequins). Each video includes two annotation types: (i) time interval and (ii) situational awareness (SA) rating. Data collection and analysis with the trials were approved by the IRB office of our research institution. 

\begin{figure*}[ht]
\vskip 0.1in
\begin{center}
\centerline{\includegraphics[width=\textwidth]{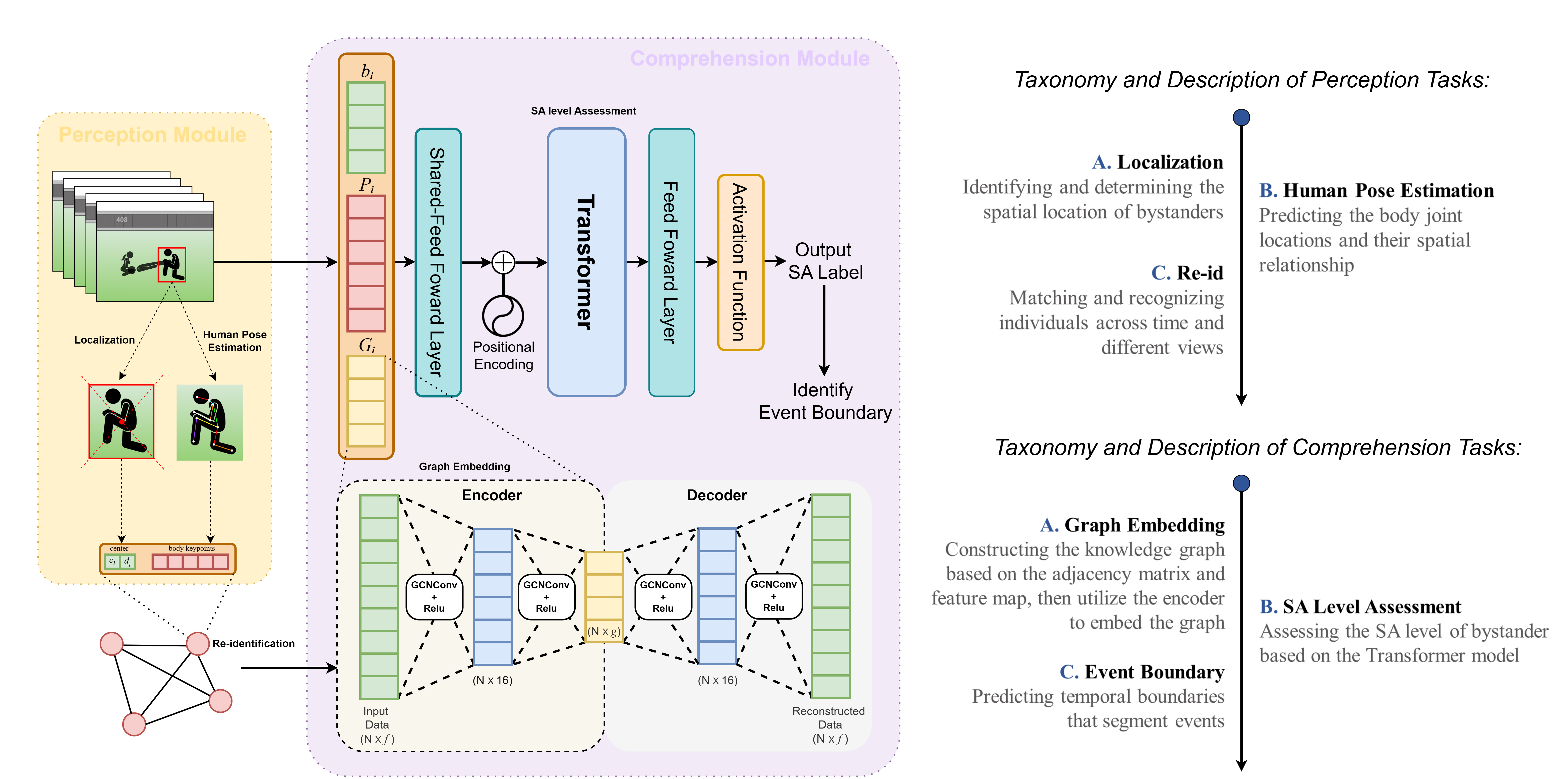}}
\caption{ Configuration for the overall perception and comprehension modules via compositional learning.}
\label{fig:frwk}
\end{center}
\vskip -0.3in
\end{figure*}

\subsection{Time Interval Annotations}


To obtain fine-grained temporal annotations, we manually segmented each video into distinct events based on the bystander's actions. To ensure consistency, two annotators independently reviewed the footage and reached consensus on event boundaries, marking specific movements or actions between events. These events span from an actor bystander's initial encounter with the simulated overdose victim (mannequin) to the successful administration of naloxone, thus capturing the entire process thoroughly.
\subsection{SA Rating Annotations}

To assess the SA of the bystanders, we first divided each video into 10 equally-sized clips, yielding 110 clips across the dataset in total. Each clip lasts 30 seconds. Two domain experts then independently reviewed each clip and rated the bystanders' SA along three dimensions: perception, comprehension, and projection, using a self-determined scale from 1 to 5. Ratings were based on observable behaviors, guided by the following questions:
\begin{itemize}
\item To what extent has the bystander observed all necessary visual cues at the current moment? \textbf {(Perception)} 
\item How well does the bystander understand the situation and their required actions at the current moment? \textbf {(Comprehension)}
\item To what degree does the bystander anticipate future developments and consequences based on the current situation? \textbf {(Projection)}
\end{itemize}

Significantly, each expert assigned a single rating to each clip only after an entire assessment, suggesting that the rating encapsulates their holistic perception of the bystander behavior throughout the duration of the segment. As such, the given SA rating reflects the bystander's SA at the final timestamp of each clip.  To enhance the richness and continuity of the training samples, we performed linear interpolation between the rated points, providing a continuous SA curve for each frame. As shown in Fig. \ref{fig:data}, The SA values are reset to 0 at event boundaries and re-evaluated thereafter to capture the dynamic nature of the SA throughout the task.
\subsubsection{SA Prediction} 
To enable the development of SA prediction models using imitation learning, we formulated the prediction problem as two classification tasks: binary and ternary classification.
\begin{itemize}
\item \textbf {Binary Classification:} Each frame is associated with a 1x3 vector representing the three aspects of SA, with each element being an integer between 1 and 5. We defined a threshold of 3, where values $\ge 3$ indicate high SA and values $<$ 3 indicate low SA. The ground truth tensor for each frame $i$ is denoted as [$Per_{i}$, $Com_{i}$, $Pro_{i}$], where $Per_{i}$, $Com_{i}$, $Pro_{i}$ $\in \left\{0, 1 \right\}$, with 1 representing high SA and 0 representing low SA.
\item \textbf {Ternary Classification:} Drawing on Endsley’s three-level model \cite{endsley1995toward}, which posits that each stage is a necessary precursor to the next higher level, higher levels of SA are meaningful only when the maximum rating at lower levels is achieved. Hence, we accumulated the binary values into a single integer for the ternary task. For instance, a binary vector of $[0, 1, 1]$ would yield a value of 0, as the perception level does not reach high SA and thus cannot contribute to higher levels. In contrast, a vector of $[1, 1, 1]$ would result in a value of 3, indicating high SA at each level and signifying a high overall SA. The resulting three classes were defined as $I \in \left\{0, 1, 2 \right\}$.

\end{itemize}

\subsection{Overall Framework via Compositional Learning}

All-in-one learning can be challenging when feature extraction and imitation learning are integrated. Here, taking advantage of compositional learning, we combined several neural networks to perform segments of the proposed SA assessment framework. By breaking down tasks into manageable units, individual networks can capitalize on their strengths for more focused processing. This approach enhances overall model performance for complex problems while maintaining reasonable computational complexity compared to the integrated learning strategies. As shown in Fig. \ref{fig:frwk},  our learning process is divided into two main phases: 1) Perception Module and 2) Comprehension Module.

\subsubsection {Perception Module}

The perception module prioritizes the interpretation of data, encompassing low-level tasks such as localization, pose estimation, and tracking. Given a sequence of video frames $\left\{ s_{1}, s_{2},...s_{l}\right\}$ with length $l$, in each frame $i$, the bounding box of the bystander is defined by a quadrilateral $b_{i}$, which contains the 2D coordinates representing the upper-left and bottom-right points of the bounding box, i.e., $\left\{x_{1}, y_{1}, x_{2}, y_{2}\right\}$. Based on the region cropped from the bounding box, 2D keypoints $P_{i} :=\left\{p_{1}, p_{2},..., p_{17} \right\}$ is  estimated for frame $i$, capturing various parts of the human body. These keypoints jointly help identify the bystander's posture and gaze. Posture implies human movements such as running, waving, or crouching, while gaze suggests their focal point, providing clues about the bystander's SA. Both tasks utilized off-the-shelf frameworks \cite{yolov8_ultralytics} trained on the Open Images V7 \cite{kuznetsova2020open} and MSCOCO dataset \cite{lin2014microsoft}, tailored for real-world applications. While a single bystander can be easily detected, confusion often arises when multiple humans are present in the scene. In such circumstances, the location identification should be assisted by maintaining unique IDs to track the objects in real-time. BoT-SORT \cite{aharon2022bot}, a state-of-the-art tracker, is used for our multi-object tracking (MOT) system.  This approach ensures robust and accurate tracking by combining object detection with re-identification techniques, allowing for consistent monitoring of each individual across frames.
     
\subsubsection {Comprehension Module} 
The comprehension module focuses on understanding and analyzing data to address higher-level tasks, such as contextual reasoning, cognition estimation, and temporal segmentation. Building upon the outputs from the perception layer and the disparity estimation, we integrated these components to model relationships between objects through graph embedding to perform contextual reasoning. We first calculated the center point of each detected bounding box, $c_{i}$, for the four relevant objects: bystander, instructor, patient, and drone. Note that the MSCOCO dataset does not include drones in its predefined categories. Therefore, we fine-tuned a pre-trained object detection model \cite{yolov8_ultralytics} using a custom dataset \cite{delleji2020deep} containing 1359 annotated drone images to capture the drone's bounding box. 

\subsubsection {Disparity Estimation} 2D coordinate-wise localization can be challenging when the real distance between objects is crucial for their interaction. For instance, in our case, the distance between the bystander and the drone might determine whether an interaction occurs. Therefore, monocular depth estimation can provide additional features to enhance understanding of their interaction. Consequently, we utilized a Dense Prediction Transformer (DPT) Model \cite{ranftl2021vision}, trained with a Vision Transformer (ViT) backbone, to provide additional features for localization. In each frame $i$, a depth label $d_{i}$ related to the center of each bounding box, $c_{i}$, was derived from the disparity estimation map $D_{i}$.

\subsubsection{Graph Embedding} The node attributes $\Phi$ of each frame $i$ is in shape $N\times f$, where $N$ represents the four different objects, and feature $f$ includes the 2D coordinates, the depth label of each center point, and the body points for each human. Zero padding is applied to the drone vector, as pose capture is not applicable. Using a fully connected graph with all edges present, represented by a $N \times N$ adjacency matrix $A$, a Graph Convolutional Network (GCN) Autoencoder is constructed to embed the interaction graph. This model learns a compact representation of the relationships among bystanders and their surroundings. The layer propagation in the graph convolution is defined as follows: 
 \begin{equation}
 f(\Phi^{(l)}, A) = \sigma(A\Phi^{(l)}W^{(l)}),
 \end{equation}
where $\Phi$ is the node attributes, $A$ is the adjacency matrix, $W$ is the learned weight matrix, $l$ is the layer, and $\sigma$ is the activation function. With a well-trained graph encoder that constructs graph representations encompassing the spatial information and poses of all humans, the embedding vector $G_{i}$ can represent the relationships of all detected objects in the scene. $G_{i}$ will be a matrix in shape $N \times g$, where $g$ is the output dimension of the last GCN layer in the graph encoder.

\subsubsection{Transformer-based Imitation Learning} 
Inspired by the TrEP model \cite{zhang2023trep}, which performs robust intention prediction of pedestrians, our transformer-based approach leverages their base architecture. We adapted it to our specific needs by individually deploying sigmoid and softmax activation functions for the binary and ternary classification tasks. We start by concatenating all extracted features $b_{i}, P_{i}, G_{i}$ at each frame $i$ to derive the feature $x_{i}$. The corresponding ground truth SA labels for binary and ternary classification are denoted as $y_{1i}$ and $y_{2i}$, respectively. The transformer module is designed to explicitly capture the temporal correlation from the input features $X_{i}={ x_{1}, x_{2}, \dots,x_{l} } $, where $l$ is the sequence length. Subsequently, the tensor $X_{i}$ is fed into the transformer model. The model first employs a shared feed-forward layer to extend the feature dimension, followed by positional encodings to inject temporal information. The core of the model comprises multiple transformer blocks. Each block contains a multi-head self-attention layer and a feed-forward layer, which transform the input vectors according to self-attention mechanisms. The resulting embeddings from the transformer blocks are then flattened and passed through a final feed-forward layer, followed by an activation function, to produce predictions of SA labels.

For the binary classification problem of predicting high or low SA, we add a sigmoid activation function at the end of the transformer model. The sigmoid function squashes the output values between 0 and 1, representing the probability of the bystander having high SA. To train the model, we use binary cross-entropy loss, which measures the difference between the predicted probabilities and the actual labels. In contrast, when the SA is categorized into three levels, we append a softmax layer to the transformer model. This layer normalizes the output values into a probability distribution over the three classes, indicating the likelihood of the bystander belonging to each SA level. During training, we use categorical cross-entropy loss, which compares the predicted probabilities to the actual one-hot encoded labels for each class.

\subsubsection{Temporal Segmentation} Leveraging the predicted SA labels and applying the rule to reset SA to 0 at the event onset, event boundaries can be identified by inferring latent SA clues from changing SA levels to divide an untrimmed video into complete actions. This approach provides insights into human SA and facilitates the discovery of connections between the bystander’s SA level alterations and event transitions.

\section{Evaluation}\label{sec:eva}

\subsection {Dataset}
We evaluate our model on the Drone-Assisted Naloxone Delivery Simulation Dataset (DANDSD). This dataset comprises 11 videos, each lasting 2-3 minutes at 50 frames per second (fps). In total, the dataset contains 92,917 frames, divided into a training set of 5,575 sequences and a testing set of 620 sequences. Each sequence is 15 frames long. During training, sequences were shuffled to enhance model learning. 
\subsection {Evaluation Metrics}

\begin{table}[htbp]
\caption{Samples achieved from each category in both Ternary and Binary Tasks.}

\label{table:Samp}
\begin{center}
\begin{tabular}{|c|c||l|c|c|}
\hline
\textbf{TERNARY} & & \textbf{BINARY} & \textbf{0} & \textbf{1} \\
\hline
0 & 34453 & Perception & 46498 & 46419 \\
\hline
1 & 33587 & Comprehension & 48838 & 44079 \\
\hline
2 & 24877 & Projection & 55885 & 37032 \\
\hline
\end{tabular}
\end{center}
\end{table}
For binary classification, we predict three SA labels per sequence; for ternary classification, a single SA label per sequence is predicted. To address the imbalance in the dataset, as shown in Table \ref{table:Samp}, we use evaluation metrics and sampling methods designed to correct for this disparity. Accuracy (Acc) with random sampling ensures balanced representation across classes. Balanced Accuracy (BAcc) averages sensitivity and specificity to account for both positive and negative classes, addressing class imbalances. The F1-Score evaluates precision and recall, providing a comprehensive measure of performance. These metrics are essential for accurately assessing changes in bystander SA across different levels of task complexity.

For temporal segmentation, we conduct the task of identifying boundaries for five predefined events, separated by specific movements agreed upon by two annotators. Evaluation metrics for segmentation include Mean over Frames (MoF) for frame-level accuracy and Intersection over Union (IoU) to assess the precision of event boundary predictions.

\begin{enumerate}
\item \textit{Mean over Frames (MoF)}:
\begin{equation}
\text{MoF} = \frac{1}{N} \sum_{i=1}^N I(y_i = \hat{y}_i),
\end{equation}
where $N$ is the total number of frames, $y_i$ is the true label for frame $i$, $\hat{y}_i$ is the predicted label for frame $i$, and $I(\cdot)$ is an indicator function.  The MoF metric can be problematic under dataset imbalance, i.e. if frequent and long action classes dominate.
\item \textit{Intersection over Union (IoU):} 
\begin{equation}
    \text{IoU} = \frac{|A \cap B|}{|A \cup B|},
\end{equation}
where $A$ is the predicted segmentation and $B$ is the ground truth segmentation.
\end{enumerate}

\begin{table}[htbp]
\caption{Ternary SA prediction performance of the proposed model and existing models on the DANDSD dataset.}
\label{table:main}
\begin{center}
\begin{tabular}{|l|c|c|c|c|}
\hline
\textbf{Model} & \textbf{Accuracy} & \textbf{AUC} & \textbf{F1} & \textbf{Precision} \\
\hline
C3D & 0.54 & 0.50 & 0.33 & 0.32 \\
\hline
I3D & 0.55 & 0.52 & 0.47 & 0.45 \\
\hline
X3D & 0.60 & 0.57 & 0.55 & 0.54 \\
\hline
SlowFast & 0.60 & 0.58 & 0.56 & 0.54 \\
\hline\hline
Ours & \textbf{0.63} & \textbf{0.64} & \textbf{0.62} & \textbf{0.62} \\
\hline
\end{tabular}
\end{center}
\end{table}

\subsection {Implementation Details}

\subsubsection{Graph Autoencoder} For node attributes $\Phi$ in shape $N\times f$,  we use a two-layer GCN that performs two propagations in the forward pass to embed $\Phi$ from $(N \times f) \rightarrow (N \times 16) \rightarrow (N \times g).$ We apply ReLU activations for each convolutional layer and use a learning rate of 0.001, training for 50 epochs.  The graph autoencoder, implemented using the PyTorch Geometric (PyG) deep learning library, trains on the same dataset as designated by DANDSD's split. 

\subsubsection{Transformer-based SA Prediction Model:} The input of the transformer-based model for SA prediction is in dimensions $(b\times l\times f)$, where $b$ refers to batch size ($b = 32$), and $l$ and $f$ refer to the sequence length ($l = 15$) and feature dimension, respectively. The initial input features are projected to 16 dimensions through the first linear layer, then expanded to 32 dimensions within the transformer's fully connected layers. There are two layers of multi-head attention (2 heads), and the dropout rates are set to 0.1. All the models are trained with Adam optimizer at a learning rate of $1e-3$ for 100 epochs. To prevent overfitting, we implemented early stopping with a tolerance of 5 epochs and employed 10-fold cross-validation to achieve better generalization of the model.

\subsection{Results}

\subsubsection{Comparison Results}
In our study, we compared the performance of several benchmark models trained on the DANDSD dataset, using a fused representation input comprising bounding box data, body keypoints, and interaction graph embeddings. The evaluated models included C3D \cite{tran2015learning}, I3D \cite{carreira2017quo}, X3D \cite{feichtenhofer2020x3d}, and SlowFast \cite{feichtenhofer2019slowfast}. Among these well-known video recognition models, C3D is a 3D convolutional network for spatiotemporal feature learning; I3D extends this by inflating 2D convolutions into 3D for enhanced information capture; X3D further optimizes efficiency by strategically expanding network dimensions; SlowFast employs a dual-pathway approach, processing video frames at different temporal resolutions. As shown in Table \ref{table:main}, our proposed model outperformed all benchmark models on the DANDSD dataset. Notably, it achieved an improvement of 3\% to 8\% across all metrics compared to the best-performing benchmark model, SlowFast. These results underscore the effectiveness of our approach in the ternary SA prediction task on the DANDSD dataset.

\subsubsection{Ablation Study Results}

\begin{table}[htbp]
\caption{Per-Class Top-1 Accuracy for each variation of the Transformer-Based Model on binary perception (Perc.), comprehension (Comp.), and projection (Proj.) prediction}
\label{table:merged}
\begin{center}
\begin{tabular}{|l|c|c|c|}
\hline
\textbf{Feature} & \textbf{Perc.} & \textbf{Comp.} & \textbf{Proj.} \\
\hline
Bbox$^{\mathrm{a}}$ & 0.70 & 0.53 & 0.62 \\
\hline
Pose$^{\mathrm{b}}$ & 0.54 & 0.52 & 0.49 \\
\hline
Graph$^{\mathrm{c}}$ & 0.54 & 0.42 & 0.37 \\
\hline
Bbox+Pose & 0.69 & 0.45 & 0.68 \\
\hline
Bbox+Graph & 0.71 & 0.45 & 0.56 \\
\hline
Pose+Graph & 0.57 & \textbf{0.61} & 0.41 \\
\hline
Bbox+Pose+Graph & \textbf{0.71} & 0.60 & \textbf{0.74} \\
\hline
\multicolumn{4}{l}{$^{\mathrm{a}}$Bounding box coordinates of the bystander.} \\
\multicolumn{4}{l}{$^{\mathrm{b}}$17 key body keypoints of the bystander.} \\
\multicolumn{4}{l}{$^{\mathrm{c}}$Interaction graph representing relationships of all 4 detected objects.}
\end{tabular}
\end{center}
\end{table}

To investigate how different features contribute to the performance of the Transformer-based model, we conducted an ablation study using various combinations of input features, including bounding boxes, body keypoints, and interaction graph embeddings. Tables \ref{table:merged} and \ref{table:sa} present the performance of the model on binary and ternary SA prediction tasks, respectively, using these different feature combinations. The results reveal two key findings. First, for both binary and ternary tasks, the fused representation combining all features achieves the best performance. Second, in both tasks, the interaction graph proves to be a valuable feature, demonstrating the relative location of each object. This potentially provides crucial clues for observing whether bystanders are paying attention to the task progress. As interactions often emerge at relatively close distances between objects, this evidence helps identify changes in events and SA. The interaction graph's effectiveness likely stems from its ability to capture spatial relationships and attention dynamics among scene participants.

\begin{table}[htbp]
\caption{Performance for each variation of the Transformer-based model on the three-level SA prediction.}
\label{table:sa}
\begin{center}
\begin{tabular}{|l|c|c|c|}
\hline
\textbf{Feature} & \textbf{Acc} & \textbf{BAcc} & \textbf{F1} \\
\hline
Bbox & 0.40 & 0.36 & 0.31 \\
\hline
Pose & 0.53 & 0.52 & 0.53 \\
\hline
Graph & 0.40 & 0.33 & 0.19 \\
\hline
Bbox+Pose & 0.43 & 0.40 & 0.38 \\
\hline
Bbox+Graph & 0.33 & 0.31 & 0.28 \\
\hline
Pose+Graph & 0.56 & 0.56 & 0.56 \\
\hline
Bbox+Pose+Graph & \textbf{0.63} & \textbf{0.62} & \textbf{0.62} \\
\hline
\end{tabular}
\end{center}
\end{table}

\subsubsection{SA curve prediction Results}
\begin{figure}[ht]
\vskip 0.2in
\begin{center}
\centerline{\includegraphics[width=1\columnwidth]{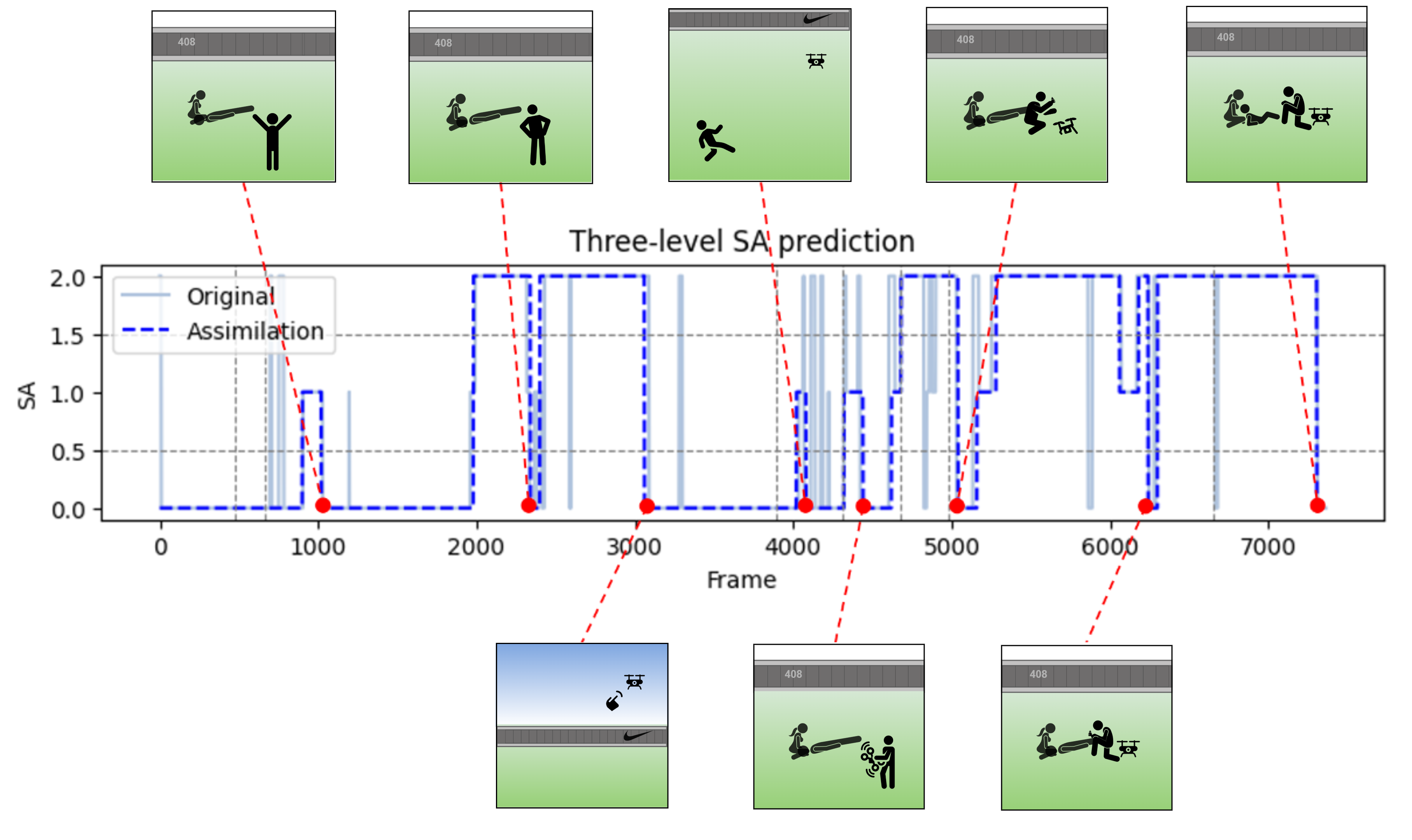}}
\caption{SA prediction performance on the testing sample. The solid line represents the SA prediction output generated by the well-trained Ternary classification model using all concatenated features as input. The dashed line depicts the output after smoothing with a 13-frame Gaussian filter, corresponding to the human reaction time of 0.25 seconds. }
\label{fig: SAC}
\end{center}
\end{figure}

Fig. \ref{fig: SAC} presents a fully delineated trajectory for a testing video sample. Considering the average human reaction time of 0.25 seconds, we track changes in the trajectory across 13 frames and apply Gaussian filtering to smooth it. Using the filtered trajectory of SA changes, we identified 8 transition points where the SA level resets to 0, effectively dividing the entire video sample into 8 segments (with the final point marking the end of the video). This approach closely mimics human cognitive processes, offering superior interpretability compared to traditional methods. The identification of transition points mirrors how humans naturally segment experiences into discrete events. Unlike frame-by-frame analyses or black-box models, our method provides insights into the evolving process of awareness that aligns with human cognition.
 
 \paragraph{Temporal Segmentation Results}

\begin{table}[htbp]
\caption{Temporal segmentation performance of transformer-based SA prediction (TrSA) and other approaches.}
\label{table:TS}
\begin{center}
\begin{tabular}{|l|c|c|}
\hline
\textbf{Method} & \textbf{MoF} & \textbf{IoU} \\
\hline
TW-Finch (cls = 6) & 0.41 & 0.23 \\
\hline
TW-Finch (cls = 7) & 0.49 & 0.29 \\
\hline
\textbf{TrSA} & \textbf{0.58} & \textbf{0.34} \\
\hline
\end{tabular}
\end{center}
\end{table}

 \begin{figure}[ht]
\vskip 0.1in
\begin{center}
\centerline{\includegraphics[width=1\columnwidth]{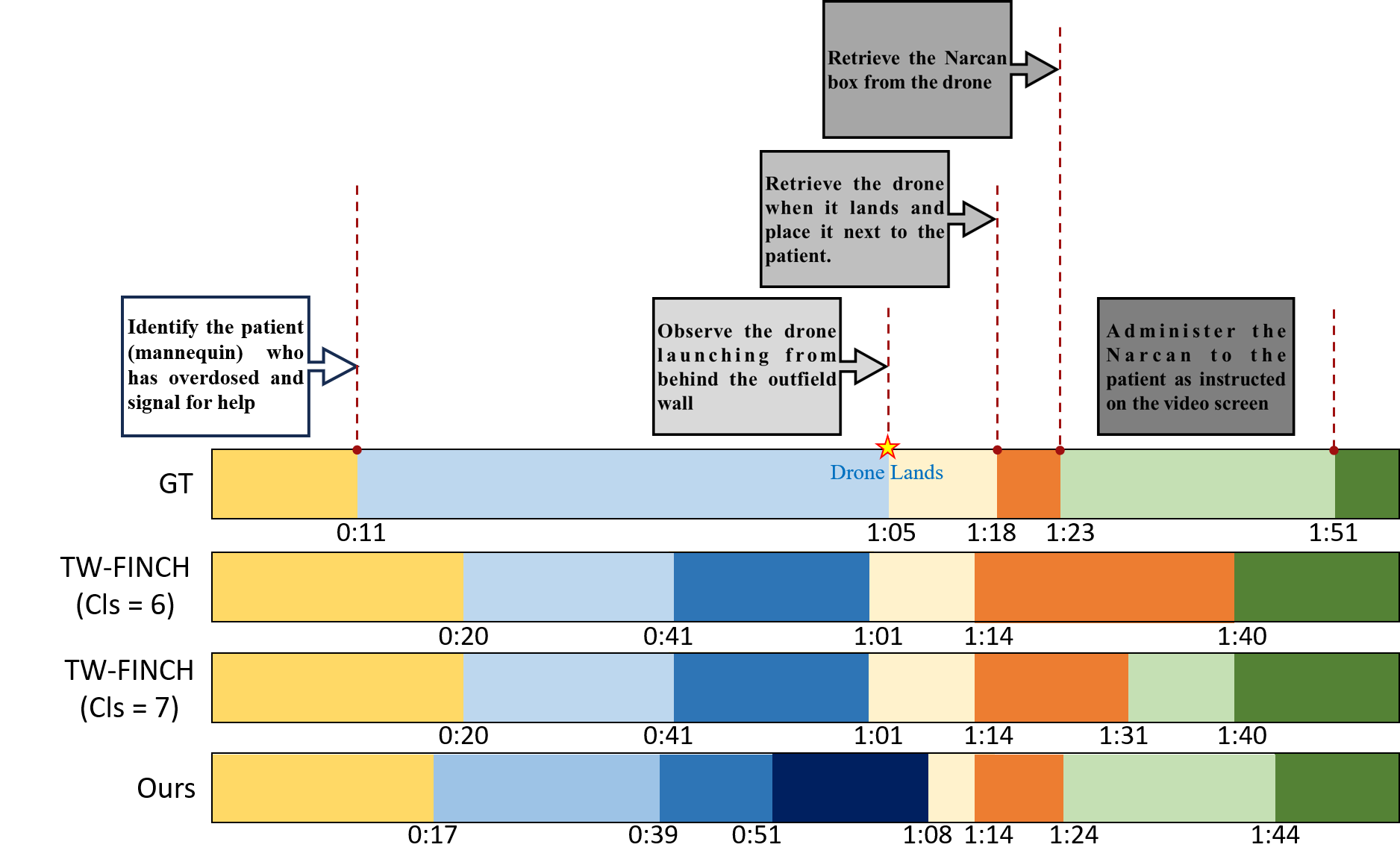}}
\vskip -0.1in
\caption{Qualitative Temporal Segmentation Results across all methods. "GT" denotes expert observations of event boundaries based on changes in bystander's SA.}
\label{fig:TSeg}
\end{center}
\end{figure}

To further explore the relationship between latent SA features and temporal segmentation (TS), we compare our model, TrSA, with the baseline TS approach, TW-FINCH \cite{sarfraz2021temporally}. Using expert-annotated event boundaries as ground truth, Table \ref{table:TS} demonstrates that TrSA surpasses TW-FINCH (cls = 7) by 9\% in Mean-over-Frames (MoF) and 5\% in Intersection-over-Union (IoU). This indicates that our approach provides a more nuanced, human-like analysis of video content. The resulting segmentation is more interpretable and cognitively aligned compared to TW-FINCH, as shown in the qualitative results in Fig. \ref{fig:TSeg}. These improvements suggest that incorporating latent SA labels significantly enhances TS performance.

\section{Conclusion}\label{sec:cls}
This research advances the field of emergency medical response by developing an AI framework for real-time situational awareness (SA) assessment in drone-assisted scenarios. Our approach, which combines graph embeddings with transformer models, offers a more comprehensive analysis of bystander behavior during simulated opioid overdose emergencies. The integration of visual, geometric, and kinematic features enables a deeper understanding of bystander-drone interactions, surpassing traditional methods in both SA prediction and temporal segmentation tasks. The significant improvements over the TW-FINCH baseline in temporal segmentation metrics highlight the robustness of our model. These advancements pave the way for more intelligent medical drone systems capable of adapting to bystander behavior in real-time. Future applications of this technology could revolutionize emergency response protocols, potentially reducing time to first intervention and improving outcomes in critical situations. As we continue to refine this approach, the implications for enhancing bystander effectiveness in emergency scenarios are substantial, offering a promising avenue for advancing AI-driven first aid systems and reducing mortality in time-sensitive medical emergencies.

\bibliographystyle{IEEEtran}
\bibliography{reference}  

\end{document}